\documentclass{article}

\usepackage{amsmath}
\usepackage{graphicx}
\usepackage{subcaption}
\usepackage{comment}
\usepackage{booktabs}
\usepackage{multirow}
\usepackage{enumitem}

\usepackage[preprint]{corl_2026} 

\newcommand{\ours}[0]{\texorpdfstring{\textbf{C}$^3$\texttt{ache}}{C³ache}}

\title{\ours: Accelerating World Action Models with Cross Inference Chunk Cache}



\author{
  Weisen Zhao\textsuperscript{1} \quad
  Lam Nguyen\textsuperscript{2} \quad
  Zhicong Lu\textsuperscript{1$\dagger$} \quad
  Yuzhang Shang\textsuperscript{2$\dagger$} \\[0.5ex]
  \textsuperscript{1}George Mason University \quad
  \textsuperscript{2}University of Central Florida \\[0.5ex]
  \texttt{\{wzhao9,zlu6\}@gmu.edu} \qquad
  \texttt{\{lam.nguyen,yuzhang.shang\}@ucf.edu}
}

\begin{document}
\maketitle
\footnotetext[0]{$\dagger$\,Corresponding authors.}


\begin{abstract}
  World Action Models (WAMs) generalize better than standard Vision-Language-Action (VLA) policies to novel motions and environments, because a video-modeling objective lets them learn from abundant unlabeled video rather than scarce labeled robot demonstrations. This generalization is computationally expensive. To complete a task, a WAM runs over multiple inference chunks, and each chunk requires a costly denoising process. Existing acceleration methods reduce this cost by caching and reusing computation within a single chunk's denoising trajectory. Our empirical analysis reveals a substantial source of redundancy they overlook: redundancy across chunks. When a robot executes a smooth behavior, the residuals computed at a given denoising step are strongly correlated from one chunk to the next. We introduce \ours, a training-free method that caches and reuses these residuals across inference chunks at the same denoising step. Experiments on benchmarks with a Fast-WAM backbone show that \ours \ achieves up to a $2.5\times$ speedup in total wall-clock inference time, with negligible degradation in task success rate. 
\end{abstract}



\section{Introduction}
\label{sec:intro}
    Vision-Language-Action (VLA) models have become the dominant recipe for generalist robot policies. VLAs such as RT-2, OpenVLA, and $\pi_0$ inherit the semantic priors of internet-scale image-text data and carry them into control~\cite{zitkovich2023rt2,kim2024openvla,black2025pi0,liu2025rdt}. However, the behaviors VLAs produce remain bounded by the demonstrations they were trained on~\cite{wam2026}. World Action Models (WAMs) come at this from the other side, pairing action prediction with a video-modeling objective that learns to predict how the scene itself will evolve~\cite{du2023unipi,wu2024gr1,hu2025vpp,wam2026}. This changes what the model can learn from: video is a dense supervisory signal, and unlike action-labeled trajectories it is available in enormous quantity. The resulting representations are shaped by how objects move and interact rather than by a fixed set of demonstrated trajectories, giving WAMs markedly stronger generalization to novel motions and unseen environments than VLAs of comparable scale~\cite{wam2026}.

\begin{figure}[t]
  \centering
  \begin{subfigure}[b]{0.66\textwidth}
    \centering
    \includegraphics[width=\linewidth]{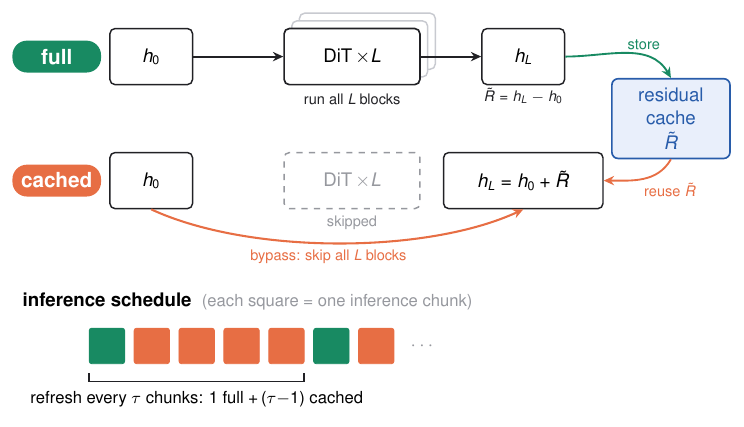}
    \caption{\ours \ method overview.}
    \label{fig:teaser}
  \end{subfigure}
  \hfill
  \begin{subfigure}[b]{0.30\textwidth}
    \centering
    \includegraphics[width=\linewidth]{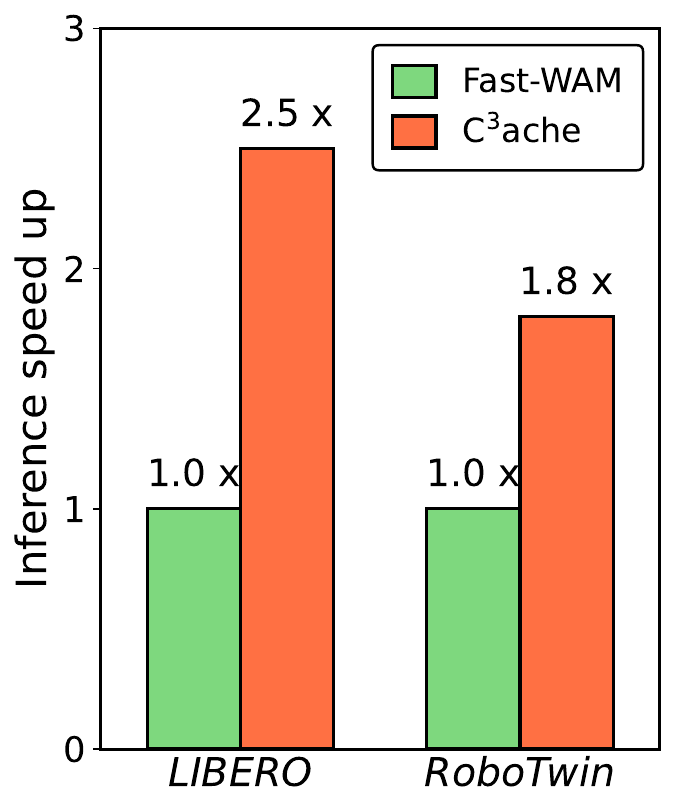}
    \caption{Inference speed-up.}
    \label{fig:speedup}
  \end{subfigure}
  \caption{(a) \textbf{Overview of \ours.} In full-computation chunks, our method runs the full DiT blocks and computes the residual. In cached chunks, it reuses the pre-computed residual to skip all DiT blocks. The residual is refreshed according to the inference schedule. (b) Inference speed comparison between \ours \ and Fast-WAM (no cache) on two benchmarks: LIBERO and RoboTwin.
}
  \label{fig:combined}
\end{figure}

    Until recently, turning this objective into a policy means following the imagine-then-execute paradigm: at test time, the model synthesizes future frames through a video-diffusion process and reads off actions from the imagined rollout. Generating frames this way is expensive. Each rollout is a long sequence of denoising steps through a diffusion transformer~\cite{peebles2023dit}, as shown in the top part of Figure~\ref{fig:teaser}, and it fits poorly with the latency budget of closed-loop control~\cite{yin2024causvid,vidarc2025}. To complete a task, a WAM runs over multiple inference chunks, and each chunk requires this expensive denoising process. This is a sequence of forward passes that each refine a noisy estimate toward a clean one~\cite{lipman2023flowmatch,black2025pi0}.
\setlength{\textfloatsep}{15pt}

    Most acceleration work targets redundancy in denoising, building on one premise: a diffusion model's features barely change between adjacent steps, so computation can be reused rather than repeated. They differ mainly in the granularity at which they cache. To accelerate visual generation diffusion model, DeepCache and Block Caching reuse whole blocks of the network across steps~\cite{ma2024deepcache,wimbauer2024blockcaching}; Learning-to-Cache instead treats each layer as the unit of reuse and learns which layers can be skipped~\cite{ma2024l2c}; ToCa pushes the granularity down to individual tokens, caching only those whose features are stable~\cite{zou2025toca}; and methods such as PAB adapt what is reused and when~\cite{zhao2025pab}. Diverse as these methods are, they share a common scope: they exploit redundancy within a single chunk's denoising trajectory (across its steps, layers, or tokens) and treat each chunk as an independent unit of computation. We therefore ask two questions: (1) Does any redundancy exist in WAM across chunks? (2) Can we exploit this redundancy to accelerate WAM's inference?


    We find that the answer to the first question is \textit{yes}: our empirical analysis reveals that this scope leaves a substantial source of redundancy untouched. When a robot executes a smooth, temporally coherent behavior, the observations at successive inference chunks change only gradually, and the action velocities predicted from them differ only slightly. We observe that this surface-level similarity propagates inward: at a fixed denoising step, the intermediate residuals computed by the action expert for one chunk are strongly correlated with those computed for the chunk before it. Caching strategies that operate one chunk at a time cannot capture this, because the quantities they would reuse never persist across the chunk.

    More importantly, we find that the answer to the second question is also \textit{yes}. We introduce \ours \ (Cross Inference Chunk Cache), a training free method that exploits precisely this cross-chunk redundancy. \ours \ caches the residuals produced at each denoising step and reuses them at the corresponding step of subsequent chunks as shown in Figure~\ref{fig:teaser}, eliminating a large fraction of the denoising computation while leaving the model weights and training procedure untouched. 


    In summary, our contributions are as follows:
    \vspace{-0.28cm}
    \begin{itemize}[leftmargin=*,noitemsep,topsep=8pt]
        \item We empirically characterize cross-chunk redundancy in a world action model, showing that the residuals computed at the same denoising step are strongly correlated across consecutive action chunks---a form of redundancy that current chunk-local caching methods do not exploit.
        \item We propose \ours, a training-free acceleration method that caches and reuses these residuals across chunks, and which composes with existing within-chunk step-wise caching.
        \item We show on LIBERO and RoboTwin~\cite{liu2023libero,mu2025robotwin}, with a Fast-WAM\cite{yuan2026fastwam} backbone, that \ours \ achieves up to a $2.5\times$ wall clock inference speedup with negligible degradation of task success rate as shown in Figure~\ref{fig:speedup}.
    \end{itemize}

\section{Related Work}
\label{sec:related}

\subsection{World Action Models}
\label{sec:related-policies}

    World Action Models pair action prediction with a video-modeling objective, learning to predict how a scene evolves~\cite{du2023unipi,wu2024gr1,hu2025vpp,wam2026}. Since the video objective can be trained on unlabeled footage, a WAM isn't bottlenecked by action-labeled robot data, and its representations are grounded in how the physical world moves. WAMs use this for imagine-then-execute inference: synthesize future frames via video diffusion, then extract actions from them. Fast-WAM~\cite{yuan2026fastwam} shows the benefit comes from video co-training, not from generating frames at inference. So a video-co-trained WAM can skip future-frame synthesis at test time and produce actions directly: a video backbone encodes the observation in one forward pass, and a flow-matching action expert outputs the action chunk. Even here, the action expert denoises each chunk over multiple steps, and this is rerun for every inference chunk.

\subsection{Caching for Diffusion Acceleration}
\label{sec:related-caching}

    Most work on speeding up diffusion models starts from one observation: a model's intermediate features change only slightly between adjacent denoising steps, so computation can be reused across steps. Methods differ in what they cache. DeepCache and Block Caching reuse whole network blocks~\cite{ma2024deepcache,wimbauer2024blockcaching}; Learning-to-Cache works at the layer level, learning which layers are safe to skip~\cite{ma2024l2c}; ToCa goes down to individual tokens, caching only the stable ones~\cite{zou2025toca}. Others adapt what is reused and when: PAB broadcasts attention outputs on a pyramid schedule, TeaCache uses the timestep embedding to decide when fresh computation is needed, and AdaCache tailors the schedule to each input~\cite{zhao2025pab,liu2025teacache,kahatapitiya2024adacache}. A related line attacks attention's quadratic cost through sparse or routed attention~\cite{sun2025vorta,zhang2025jenga} or post-training compression, as in DiTFastAttn~\cite{yuan2024ditfastattn}. All of these share one scope: they exploit redundancy inside a single inference chunk, treating each chunk as independent. In contrast, our method leverages redundancy across inference chunks.

\subsection{Accelerating Iterative Action Generation}
\label{sec:related-action}
     One way to cut the cost of iterative action generation is to shorten the schedule. Distillation compresses a many-step denoiser into a few step or single-step one; SnapFlow, for instance, distills a flow-matching action expert down to a single forward pass~\cite{luan2026snapflow}. These methods reduce the number of steps, but still treat each chunk's generation in isolation.

    The second axis of redundancy comes from the way long behaviors are produced: chunk by chunk, in sequence. Autoregressive video diffusion makes this explicit: a long video is generated chunk by chunk, each new segment denoised conditioned on the frames already produced~\cite{chen2024diffusionforcing,song2025dfot}. Because each chunk attends to earlier ones, key-value caching lets a new chunk reuse attention states already computed~\cite{huang2025selfforcing}, and Ca2-VDM precomputes and shares them across denoising steps~\cite{gao2025ca2vdm}. CausVid takes a different approach, distilling a slow bidirectional video diffusion model into a fast autoregressive one~\cite{yin2024causvid}.


    The closest work to ours is X Cache~\cite{zeng2026xcache}, concurrent work that also identifies cross-chunk redundancy: the block inputs at matching denoising-step and block positions are similar between consecutive chunks. X-Cache is built for few-step distilled driving world models, where inter-step redundancy has already been removed by distillation. In contrast, \ours \ targets the standard multi-step denoising regime, where this redundancy remains to be exploited. In short, X-Cache removes cross-chunk redundancy in a setting where inter-step redundancy is already gone, whereas \ours \ targets the standard multi-step setting and exploits both axes together.


\section{Method}

This section outlines the proposed \ours \ framework. We begin in Section~\ref{sec: prelim} by reviewing the underlying formulations our method builds upon. Section~\ref{sec:key-observation} then highlights the key empirical observations that motivate our approach. Finally, we break down the detailed workflow of \ours \ in Section~\ref{sec: method_design}, with a visual overview provided in Figure~\ref{fig:work_flow}.
\subsection{Preliminaries}
\label{sec: prelim}    
\textbf{World action models with multi-chunk action generation.}
\hspace{0.3cm}Actions are produced in chunks. At control cycle $c$ the action expert outputs a chunk $x^c$ of $H$ future  actions; the robot executes part of that chunk, the scene advances, and the next cycle conditions on a  fresh observation to produce $x^{c+1}$. A full episode is thus a sequence of chunks $x^1, x^2, \dots$, each generated from its own context $\mathrm{ctx}^c$. We refer to each $x^c$ as an inference chunk. 

\textbf{Action generation by flow matching.}\hspace{0.3cm}A chunk is not produced in a single network evaluation. The action expert is a flow-matching model~\cite{lipman2023flowmatch,liu2023rectifiedflow}: it turns a sample of noise into a clean action chunk through a sequence of denoising steps. We use the flow matching formulation throughout.

    Flow matching connects the clean action chunk $x^c$ to a Gaussian sample
    $\epsilon^c \sim \mathcal{N}(0, I)$ along the straight path
    \begin{equation}
        x^c_t = (1-t)\,\epsilon^c + t,x^c,
        \qquad t \in [0,1],
        \label{eq:fm-path}
    \end{equation}
    so that $t = 0$ is pure noise and $t = 1$ is the action chunk. The action expert $v_\theta$ is trained to predict the velocity of this path, $x^c - \epsilon^c$, given the noisy iterate, the time $t$, and the context $\mathrm{ctx}^c$. At inference the chunk is generated by integrating the resulting ODE from noise to data. Discretizing $[0,1]$ into $K$ steps $0 = t_0 < t_1 < \dots < t_K = 1$ and writing $x^c_{(k)}$ for the iterate after step $k$, the sampler starts from $x^c_{(0)} = \epsilon^c$ and repeats
    \begin{equation}
        x^c_{(k)} = x^c_{(k-1)} + (t_k - t_{k-1})\, v^c_k,
        \qquad
        v^c_k = v_\theta\!\left(x^c_{(k-1)},\, t_{k-1},\, \mathrm{ctx}^c\right),
        \label{eq:fm-step}
    \end{equation}
    for $k = 1, \dots, K$, after which $x^c_{(K)}$ is the action chunk the robot executes. We call $v^c_k$ the velocity prediction at denoising step $k$ of chunk $c$.

    The cost of an episode follows directly. Producing one action chunk takes $K$ denoising steps of the action expert. An episode of $C$ control cycles takes $C \cdot K$ of them in all. Every one is a full forward pass of the action expert, and chunk $c$ is denoised from fresh noise $\epsilon^c$ with no reference to the computation already spent on previous chunks. Our method \ours \ changes that.

\textbf{The action expert as a diffusion transformer.}\hspace{0.3cm}
    We now look more closely at the action expert. The noisy action chunk is embedded into a sequence of action tokens, which we write $h_0$; the denoising time $t$ enters through adaptive normalization. These tokens pass through a stack of $L$ Diffusion Transformer (DiT) blocks, each pairing an attention layer with a feedforward network, with every sublayer wrapped in a residual connection. The attention is the joint attention of the MoT~\cite{liang2025mixtureoftransformers}: an action token attends both to the other action tokens and to the video tokens that carry the conditioning $\mathrm{ctx}^c$. Within a chunk those video tokens are fixed, since the video branch produces them once, so over the $K$ denoising steps the action expert's computation varies only with the action tokens. Writing $h_L$ for the output of the final block, the velocity is read from $h_L$ by a linear head.
    
    Because every block carries a residual connection, the input $h_0$ propagates additively through the entire stack: the output is $h_L = h_0 + R$, where $R$ collects everything the $L$ blocks add. We call $R$ the residual of the action expert,
    \begin{equation}
        R = h_L - h_0,
    \end{equation}
    the output of the last block minus the input to the first. A single forward pass of the action expert produces one such $R$. Two facts about it matter here. First, producing $R$ is the whole cost: it means running all $L$ blocks---their joint attention, whose cost grows quadratically with the number of tokens~\cite{sun2025vorta,zhang2025jenga}, and their feedforward networks---which is essentially the entire forward pass. Second, $R$ is a self-contained additive term: given $R$, the output is recovered as $h_L = h_0 + R$ and the velocity follows, with no block evaluated. An estimate of $R$ obtained without running the stack would therefore let a denoising step be taken almost for free.

    Indexed by chunk and step, write $h^{c,k}_0$ and $h^{c,k}_L$ for the input embedding and stack output at denoising step $k$ of chunk $c$, and $R^c_k = h^{c,k}_L - h^{c,k}_0$ for the residual. The output head turns the stack output into the velocity prediction of that step,
    \vspace{-0.8em}
    \begin{equation}
        v^c_k \;=\; \mathrm{Head}\bigl(h^{c,k}_L\bigr)
                \;=\; \mathrm{Head}\bigl(h^{c,k}_0 + R^c_k\bigr),
        \label{eq:velocity}
    \end{equation}
    \vspace{-2.2em}

     \ours \ rests on the observation, developed in the next section, that across the chunk index $c$, at a matched step $k$, these residuals have high similarity across inference chunks.

\subsection{Key Observation}
\label{sec:key-observation}
    To investigate what information can be reused across inference chunks, we perform our analysis using Fast-WAM~\cite{yuan2026fastwam} on the LIBERO~\cite{liu2023libero} benchmark. LIBERO contains four suites: LIBERO Spatial, LIBERO-Object, LIBERO-Goal, and LIBERO-Long. Each suite contains 500 episodes covering 10 tasks. We implement the analysis with $2,000$ episodes containing all suites.

    \begin{figure}[t]
        \centering
        \includegraphics[width=0.92\linewidth]{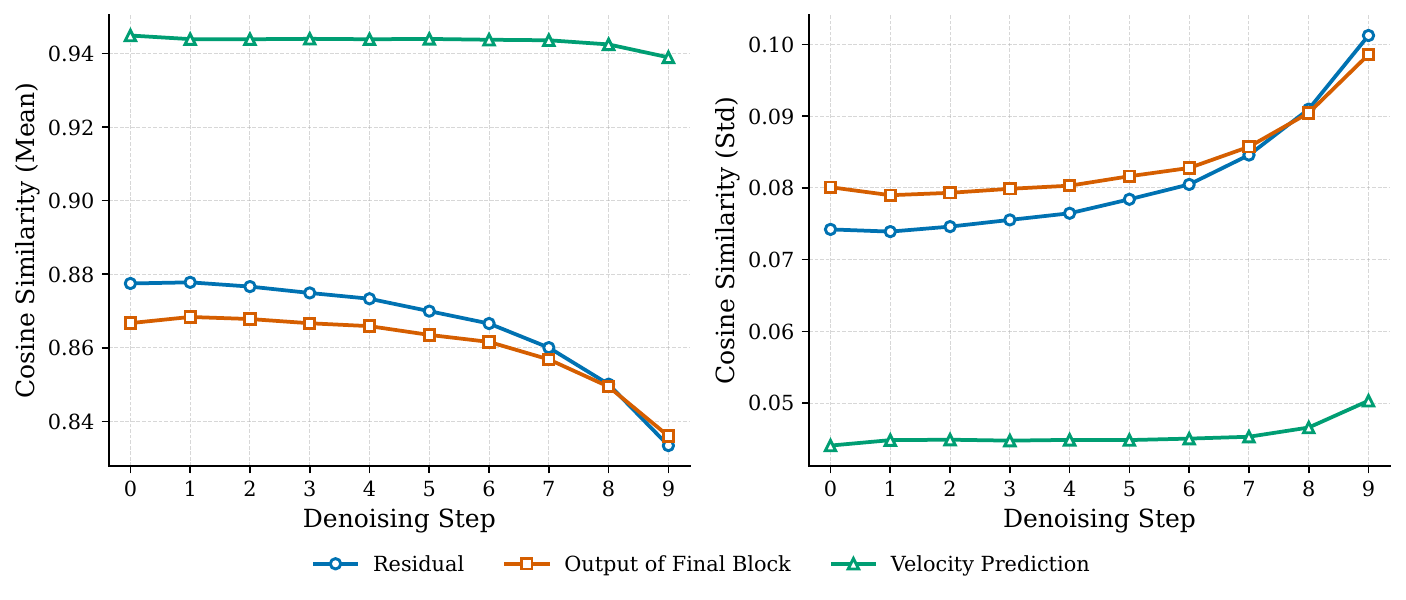}
        \caption{Cosine similarity across inference chunks for the three tensors: output of Final Block $h_L$, Residual $R$ and Velocity Prediction $v$. Left: mean; right: standard deviation.}
        \label{fig:cos_sim}
    \end{figure}
    
    We examine three tensors: the output of the final block $h_L$, the residual $R$, and the velocity prediction $v$. Specifically, we compute the cosine similarity of $h_L$, $R$, and $v$ across inference chunks at the same denoising step, using 10 denoising steps per chunk over the full LIBERO benchmark:
    \begin{equation}
        \resizebox{0.93\textwidth}{!}{$
        \text{sim}(h^{c-1,k}_L, h^{c,k}_L) = \frac{h^{c-1,k}_L \cdot h^{c,k}_L}{\|h^{c-1,k}_L\| \, \|h^{c,k}_L\|}, \quad
        \text{sim}(R^{c-1}_k, R^c_k) = \frac{R^{c-1}_k \cdot R^c_k}{\|R^{c-1}_k\| \, \|R^c_k\|}, \quad
        \text{sim}(v^{c-1}_k, v^c_k) = \frac{v^{c-1}_k \cdot v^c_k}{\|v^{c-1}_k\| \, \|v^c_k\|}
        $}
    \end{equation}
    The evaluation on the whole LIBERO benchmark yields $29,924$ inference chunks. At each denoising step, we report the mean and standard deviation of the cosine similarity across all $29,924$ chunks. As shown in Figure~\ref{fig:cos_sim} (left), adjacent inference chunks exhibit very high cosine similarity at the same denoising step, revealing substantial redundancy that can potentially be exploited to reduce computation. In addition, as shown in Figure~\ref{fig:cos_sim} (right), the standard deviation of cosine similarity increases as the denoising process approaches its end, indicating greater variability across inference chunks at the tail denoising steps. This suggests that, when caching, the last few denoising steps should be avoided, a choice that is confirmed by our later experiments in Section~\ref{sec:experiment}. 

\begin{figure}[t]
  \centering
  \includegraphics[width=0.92\linewidth]{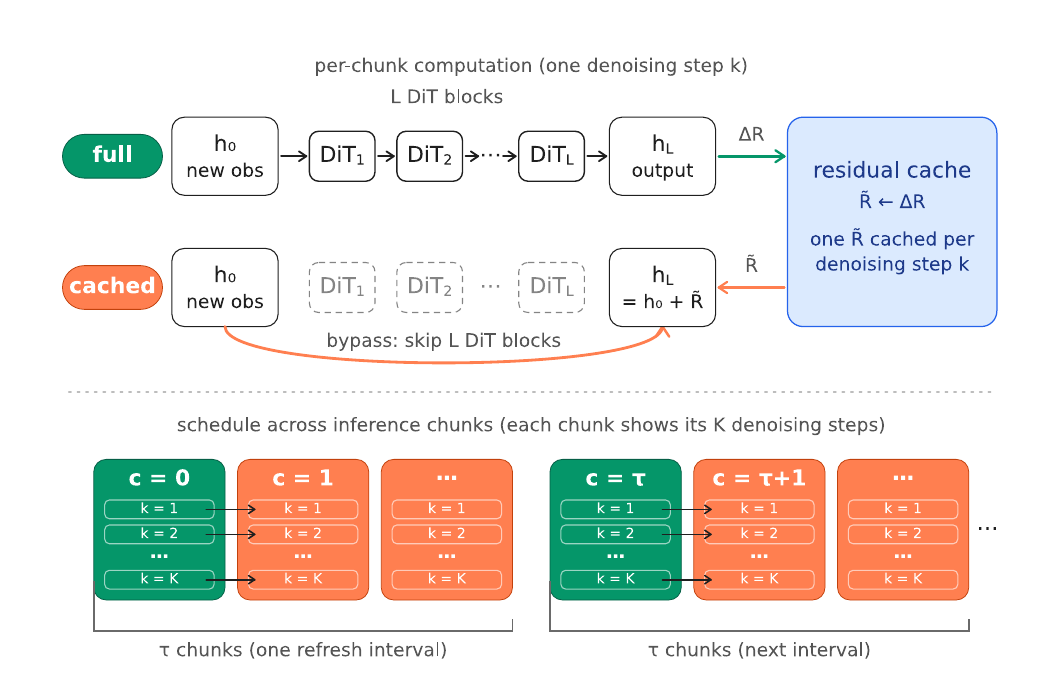}
  \caption{%
    \textbf{\ours \ framework design.} \textbf{(Top)} Per-chunk computation at one denoising step~$k$. A \emph{full} chunk runs all $L$ DiT blocks and writes the residual $\Delta R = h_L - h_0$ into the cache as $\tilde{R}$; a \emph{cached} chunk skips every DiT block and reconstructs $h_L = h_0 + \tilde{R}$ from the current observation~$h_0$ and the cached residual. \textbf{(Bottom)} Schedule across inference chunks. Chunks are grouped into refresh intervals of length~$\tau$: every $\tau$-th chunk is fully recomputed and refreshes the cache, while every other chunk in the interval reads from it; the first chunk is always full computation. A separate residual~$\tilde{R}$ is cached per step.%
  }
  \label{fig:work_flow}
\end{figure}

\subsection{\ours}
\label{sec: method_design}
    As shown in Section~\ref{sec:key-observation}, $h_L$, $R$, and $v$ at the same denoising step are highly correlated across inference chunks. They vary smoothly across inference iterations in this autoregressive action generation setup.  We consider caching those three tensors to reuse this redundancy to speed up action inference. Directly caching $h_L$ and $v$ freezes the two tensors at each denoising step across consecutive inference chunks. Therefore, $h_L$ and $v$ are excluded from cache choices. To allow the new observation to flow into the current chunk while still reusing the redundancy from previous inference chunks, we ultimately choose to cache $R$.
    
    When caching is off and $h_L$ is fully computed at denoising step $k$ and at inference chunk $c$, we cache the residual as follow:
    \vspace{-0.3em}
    \begin{equation}
        \tilde{R}_{c,k}\gets\Delta R,
        \qquad
        \Delta R = h^{c,k}_L - h^{c,k}_0,
        \label{eq:residual}
    \end{equation}
    \vspace{-0.3em}
    In the next inference chunk $c'$, at denoising step $k'$ with caching on, we combine $h^{c',k'}_0$ with $\tilde{R}_{c,k}$ to skip the expensive computation of the $L$ DiT blocks, since $h^{c',k'}_0$ already carries the new observation from chunk $c'$:
    \vspace{-0.3em}
    \begin{equation}
        h^{c',k'}_L= h^{c',k'}_0 + \tilde{R}_{c,k}
    \end{equation}
    
    We define a cache refresh interval $\tau$. Concretely, $\tilde{R}_{c,k}$ is reused to compute the block output $h_L$ for chunks $c$ through $c+\tau-1$, and we refresh the cache to $\tilde{R}_{c+\tau,k}$ at chunk $c+\tau$. Caching with reuse cannot be applied in the first chunk, since $\tilde{R}_{0,k}$ is empty when $c=0$; in this case we run the full denoising computation to populate $\tilde{R}_{0,k}$. Without such a refresh, the cached residual would stay frozen at $\tilde{R}_{0,k}$ throughout inference, which is the case $\tau = 0$.
    \begin{equation}
        \tilde{R}_{c,k} =
            \begin{cases}
                \Delta R, & c \bmod \tau = 0 \\[4pt]
                \tilde{R}_{c-1,k}, & \text{otherwise}
            \end{cases}
        \label{eq:cache-update}
    \end{equation}
Figure~\ref{fig:work_flow} illustrates the detailed design of \ours{} and how it operates at inference time across inference chunks.


\section{Experiment}
\label{sec:experiment}
\subsection{Experimental Settings}
\textbf{Model and Datasets.}\hspace{0.3cm}
    We evaluate our method on two benchmarks: LIBERO~\cite{liu2023libero} and RoboTwin 2.0~\cite{mu2025robotwin}. For LIBERO, we evaluate on all four task suites, which together contain 2000 episodes, as discussed in Section~\ref{sec:key-observation}. RoboTwin 2.0 is a challenging bimanual manipulation benchmark consisting of 50 tasks that require coordinated dual-arm control. We evaluate on all 50 tasks under two settings: \emph{clean} and \emph{randomized}. In the clean setting, object poses are randomized but the environment is left unchanged. In the randomized setting, it additionally varies the background, lighting, table height, camera distance, and clutter. For each task, we run 20 evaluation episodes per setting, each with a different seed. In the clean setting, varying the seed changes the initial object placement while the environment is held fixed. In the randomized setting, it changes both the initial object placement and the environment, including all visual conditions.

    We evaluate our method on Fast-WAM~\cite{yuan2026fastwam}, which builds on the Wan2.2-5B backbone~\cite{wan2025}. Fast-WAM keeps video co-training during training but skips future prediction at action inference time. Its action expert mirrors the architecture of the video branch, but with a reduced hidden dimension of $d_a = 1024$. This gives a 1B-parameter action expert and a total model size of 6B parameters. For both LIBERO and RoboTwin, we use 10 denoising steps and an action horizon of 32.
    
\textbf{Hardware.}\hspace{0.3cm}
We evaluate our method on a single NVIDIA A100 80GB GPU with a 64-Core AMD CPU and 500 GB memory by PyTorch for all experiments.

\textbf{Metrics.}\hspace{0.3cm}
We evaluate whether our method improves inference speed without noticeably degrading performance. We report two metrics: total wall-clock inference speedup across all episodes and success rate. The reported inference speed covers the full inference pipeline, not just the cached DiT blocks: it also includes the components we never cache, namely VAE encoding, video prefill for KV cache construction, action token embedding, the projection from block outputs to velocity predictions, and scheduler steps.

\textbf{Implementation Detail.}\hspace{0.3cm}
For both LIBERO and RoboTwin, we use 10 denoising steps and an action horizon of 32. As shown in Figure~\ref{fig:cos_sim} and discussed in Section~\ref{sec:key-observation}, both the mean and standard deviation of the cosine similarity across inference chunks drop more sharply past step 6. We therefore evaluate our method on LIBERO and RoboTwin across two cache-step ranges, $[0,6]$ and $[0,7]$, each paired with refresh intervals $\tau \in \{0, 4, 8\}$. For the ablation, we repeat the same $\tau$ values with longer ranges, $[0,8]$ and $[0,9]$.

\subsection{Main Results}
Table~\ref{tab:libero_matrix} and Table~\ref{tab:robotwin_matrix} summarize the results on LIBERO and RoboTwin, respectively. Overall, \ours{} achieves wall-clock inference speedups across the cache step ranges $[0,6]$ and $[0,7]$ with negligible degradation in task success rate on both LIBERO and RoboTwin. The best speedups we obtain are $2.5\times$ on LIBERO and $1.8\times$ on RoboTwin. At $2.5\times$, LIBERO shows no drop in success rate, and at $1.8\times$, RoboTwin loses only $1\%$. Within this range, RoboTwin also reaches $1.6\times$ speedup with no drop at all. In addition, we find that more frequent cache refreshes are needed in the cache step range $[0,7]$ on RoboTwin. Looking into why, we find a clear trend with task length: at $\tau=0$, all shortest tasks (3--5 chunks per episode) hold their success rates with no drop, tasks of moderate length (6--11 chunks) drop by $10.81\%$ on average, and tasks with 12 or more chunks fall by $34\%$---a substantial degradation. This monotonic trend suggests that longer-horizon episodes, where stale cache compounds over more steps, call for a lower $\tau$.

\begin{table}[t]
\centering
\caption{\textbf{Results on LIBERO.} Effect of cache steps and refresh interval on success rate and inference speedup. SR is reported per suite (\%), with $\Delta$SR measured against the baseline (Fast-WAM). Speedups are relative to the baseline (Fast-WAM).}
\label{tab:libero_matrix}
\setlength{\tabcolsep}{4.9pt}
\begin{tabular}{lccccccccc}
\toprule
& Cache & & \multicolumn{5}{c}{Success Rate (\%)} & & \\
\cmidrule(lr){4-8}
Config & Steps & Refresh & Spatial & Object & Goal & Long & Average & $\Delta$SR & Speedup \\
\midrule
Fast-WAM & --- & --- & 96.40 & 99.40 & 96.40 & 95.40 & 96.90 & $+0.00$ & $1.00\times$ \\
\midrule
\multirow{3}{*}{$[0,6]$}
 & 7 & 0 & 97.40 & 100.00 & 97.20 & 95.20 & 97.45 & $+0.55$ & $2.10\times$ \\
 & 7 & 4 & 97.80 & 100.00 & 95.60 & 92.80 & 96.55 & $-0.35$ & $1.69\times$ \\
 & 7 & 8 & 98.00 & 99.80 & 95.40 & 93.80 & 96.75 & $-0.15$ & $1.86\times$ \\
\midrule
\multirow{3}{*}{$[0,7]$}
 & 8 & 0 & 97.60 & 99.60 & 96.00 & 95.20 & \textbf{97.10} & $\mathbf{+0.20}$ & $\mathbf{2.51\times}$ \\
 & 8 & 4 & 96.80 & 99.80 & 95.60 & 94.60 & 96.70 & $-0.20$ & $1.82\times$ \\
 & 8 & 8 & 97.40 & 99.80 & 96.40 & 94.40 & 97.00 & $+0.10$ & $2.13\times$ \\
\bottomrule
\end{tabular}
\end{table}

\begin{table}[t]
\centering
\caption{\textbf{Results on RoboTwin.} Effect of cache steps and refresh interval on success rate and inference speedup. SR is reported on clean and randomized settings (\%), with $\Delta$SR measured against the baseline (Fast-WAM). Speedups are relative to the baseline (Fast-WAM).}
\label{tab:robotwin_matrix}
\begin{tabular}{lccccccc}
\toprule
& Cache & & \multicolumn{3}{c}{Success Rate (\%)} & & \\
\cmidrule(lr){4-6}
Config & Steps & Refresh & Clean & Random & Average & $\Delta$SR & Speedup \\
\midrule
Fast-WAM & --- & --- & 92.20 & 90.60 & 91.40 & $+0.00$ & $1.00\times$ \\
\midrule
\multirow{3}{*}{$[0,6]$}
 & 7 & 0 & 91.00 & 89.70 & 90.35 & $-1.05$  & $\mathbf{1.84\times}$ \\
 & 7 & 4 & 92.60 & 91.10 & \textbf{91.85} & $\mathbf{+0.45}$ & $1.56\times$ \\
 & 7 & 8 & 90.90 & 89.90 & 90.40 & $-1.00$ & $1.71\times$ \\
\midrule
\multirow{3}{*}{$[0,7]$}
 & 8 & 0 & 76.10 & 75.60 & 75.85 & $-15.55$ & $1.68\times$ \\
 & 8 & 4 & 88.60 & 89.20 & 88.90 & $-2.50$  & $1.55\times$ \\
 & 8 & 8 & 85.40 & 85.10 & 85.25 & $-6.15$  & $1.72\times$ \\
\bottomrule
\end{tabular}
\end{table}

\begin{figure}[t]
  \centering
  \begin{subfigure}[t]{0.49\linewidth}
    \includegraphics[width=0.86\linewidth]{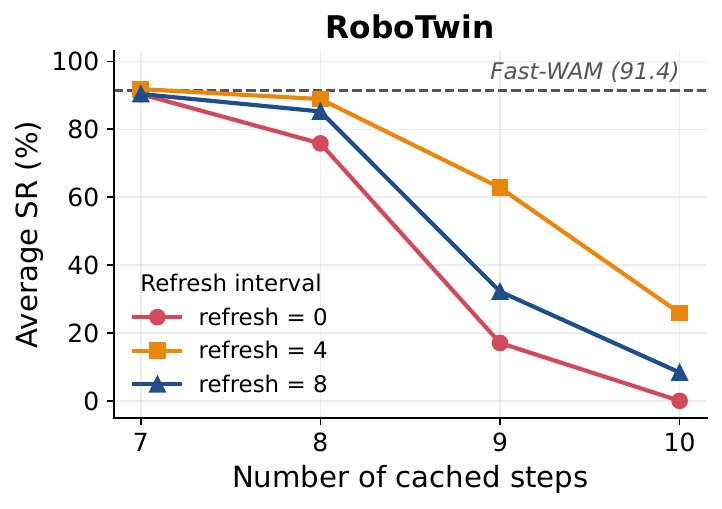}
    \caption{RoboTwin}
    \label{fig:sr_robotwin}
  \end{subfigure}
  \hfill
  \begin{subfigure}[t]{0.49\linewidth}
    \includegraphics[width=0.8\linewidth]{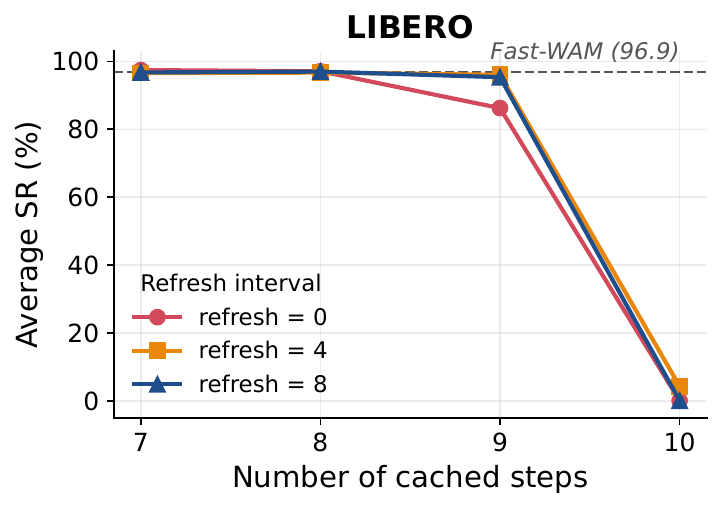}
    \caption{LIBERO}
    \label{fig:sr_libero}
  \end{subfigure}
  \caption{\textbf{Effect of cache-step range on success rate.} On both benchmarks, longer
           caching steps causes a sharp drop in success rate}
  \label{fig:sr_vs_cache_depth}
  \vspace{-0.15in}
\end{figure}

\subsection{Ablation}
\label{sec:ablation}
We keep the same $\tau$ values with longer cache-step ranges, $[0,8]$ and $[0,9]$. As shown in Figure~\ref{fig:sr_vs_cache_depth}, performance falls off sharply in both cases. On LIBERO, the task success rate drops substantially at $\tau = 0$ for the range $[0,8]$, and at every $\tau$ for the range $[0,9]$; on RoboTwin, it drops at every refresh interval $\tau$ for both ranges. This is consistent with our observation in Section~\ref{sec:key-observation} that the tail denoising steps vary more across inference chunks.

\vspace{-0.12in}
\section{Limitation}
    We find that the refresh interval $\tau$ is an important factor affecting both inference speedup and success rate. For example, in our RoboTwin experiments, we observe that within the cache step range $[0,7]$ setting, an appropriate refresh interval $\tau$ can deliver a decent inference speedup while avoiding excessive loss in success rate. In general, a smaller and more conservative $\tau$ leads to a relatively small drop in success rate but also yields a lower inference speedup, whereas a larger and more aggressive $\tau$ sacrifices more success rate while achieving a relatively higher inference speedup. In the current design, the refresh interval $\tau$ must be tuned manually to find the most suitable balance point. In future work, we hope to find a dynamic refresh interval based on task difficulty, the complexity of scene, and the rate of change of the physical situation, so as to reach the sweet spot between success rate and inference speedup---that is, how to maximize inference speedup while incurring little loss in success rate.
    \vspace{-0.15in}


\section{Conclusion}
\vspace{-0.15in}
In this paper, we propose \ours, a training-free method that caches and reuses residuals across inference chunks at the same denoising step, reducing the total inference time of WAM with negligible drop in task success rate. \ours\ achieves up to a $2.5\times$ speedup in wall-clock inference time while maintaining task success rate on LIBERO and RoboTwin.

\bibliography{citation_corl}  

\end{document}